# Boosting of Classification Models with Human-in-the-Loop Computational Visual Knowledge Discovery

Alice Williams[0009-0001-6154-2407] and Boris Kovalerchuk[0000-0002-0995-9539]

Department of Computer Science, Central Washington University, USA
{williamscade, borisk}@cwu.edu

**Abstract.** High-risk artificial intelligence and machine learning classification tasks, such as healthcare diagnosis, require accurate and interpretable prediction models. However, classifier algorithms typically sacrifice individual case-accuracy for overall model accuracy, limiting analysis of class overlap areas regardless of task significance. The Adaptive Boosting meta-algorithm, which won the 2003 Gödel Prize, analytically assigns higher weights to misclassified cases to reclassify. However, it relies on weaker base classifiers that are iteratively strengthened, limiting improvements from base classifiers. Combining visual and computational approaches enables selecting stronger base classifiers before boosting. This paper proposes moving boosting methodology from focusing on only misclassified cases to all cases in the class overlap areas using Computational and Interactive Visual Learning (CIVL) with a Human-in-the-Loop. It builds classifiers in lossless visualizations integrating human domain expertise and visual insights. A Divide & Classify process splits cases to simple and complex, classifying these individually through computational analysis and data visualization with lossless visualization spaces of Parallel Coordinates or other General Line Coordinates. After finding pure and overlap class areas simple cases in pure areas are classified, generating interpretable sub-models like decision rules in Propositional and First-order Logics. Only multidimensional cases in the overlap areas are losslessly visualized simplifying end-user cognitive tasks to identify difficult case patterns, including engineering features to form new classifiable patterns. Demonstration shows a perfectly accurate and losslessly interpretable model of the Iris dataset, and simulated data shows generalized benefits to accuracy and interpretability of models, increasing end-user confidence in discovered models.

## 1 Introduction

Improving both accuracy and explainability of classification learning models is a fundamental challenge in machine learning (ML). The Adaptive Boosting (AdaBoost) meta-algorithm [4, 13] which won the 2003 Gödel Prize improves accuracy of classification models analytically by assigning higher importance weights to misclassified cases yielded by the weaker classifiers to run classifiers on again. Boosting algorithms, including AdaBoost, are popular in both data science and ML competitions where many winners use boosting algorithms to achieve high accuracy [13].

The goal of this paper is to enhance ML boosting methods by shifting focus from just the boosting of the misclassified cases to the boosting of all the class overlap areas, by combining Computational and Interactive Visual Learning (CIVL) with a Human-in-the-loop. Considering all overlap areas of data classes, beyond only the misclassified cases, is needed to achieve higher ML model confidence and trust, especially for high-stakes ML tasks required from domain experts/end users.

Shifting focus from just the misclassified cases to all the overlap areas was motivated by the well-established importance of the class overlap areas [11, 14, 15, 17, 18]. Areas where a mixture of cases from different classes are present are difficult to distinguish [2, 14, 15, 17, 18]. Many methods have been previously proposed to define and handle ML class overlap, such as fuzzy-logic and probabilistic Bayesian models. The conceptual diagram in [11] presents a version of a separating scheme approach to separate pure and overlap areas in ML. There is a deep link between the class overlap and model explainability/transparency challenges [1] to provide explanations including counterfactuals when cases of the opposite classes are nearby. The focus of model boosting on the class overlap areas allows a significant decrease in the amount of data requiring exploration, which helps to avoid inflated ML model accuracy estimates.

The computational part of CIVL is based on traditional computational ML algorithms and models, like linear and nonlinear classifiers: decision trees (DT), logical propositional and first order logic rules, shallow and deep neural networks, and others. The visual learning part is based on the Parallel Coordinates, and other General Line Coordinates (GLC), which can losslessly visualize high-dimensional tabular data [7, 8, 10]. This visual learning approach is a part of the Interpretable Computational Visual Knowledge Discovery (VKD) methodology [9, 10].

Visualization methods help with understanding the nature and extent of class overlap, which is crucial for assessing the difficulty of classification problems and the performance of ML algorithms. Several methods are already available to visualize classes and overlap area as convex hulls. However, in high-dimensional spaces computing exact convex hulls becomes computationally expensive [3], and often infeasible. This motivated *contour plots* and *bounding boxes*, where instead of using strict convex hull, contour plots show a generalized "shape" of each class. Existing visualization methods do not fully visualize high-dimensional convex hulls for dimensions above 3D. Parallel Coordinates and other GLCs have emerged as a powerful tool for visualizing multidimensional data without loss of information [5-11] which we use in this paper. It is shown that the shape of the convex hull in 2D Parallel Coordinates is more complex than in 2D Cartesian Coordinates [8]. Therefore, we propose a simpler version of a convex hull in multidimensional (n-D) Parallel Coordinates as a *bounding box*.

The key idea of the proposed CIVL framework is separating areas of feature space, where the models can be discovered relatively easily from much smaller sub-areas, instead of focusing on the full feature space, which would require a different approach and significantly more effort. This is a version of the classical Divide and Conquer strategy, which we call a *Divide and Classify* (D&C) strategy, or a *Separating Scheme* for searching of pure and overlap areas of classes and building separate models in these areas until all cases will be classified preferably with interpretable models/rules.

Several studies demonstrated the benefits of building ML models directly on the overlapped regions and separately from the non-overlapped regions by using the separating scheme. For instance, the approach in [14] first uses k-NN to extract overlapping regions, then trains two separate SVMs for the overlapping and non-overlapping regions. A related idea from [16] produced an interpretable ML model, where data areas are divided to two categories: (1) areas where an interpretable (white-box) model can be and is generated and (2) areas where an uninterpretable (black-box) model is generated, since it is difficult to generate an interpretable model for that area. We also successfully used this scheme in our synthetic data generation and labeling algorithm [19].

The proposed CIVL framework consists of several key steps: (a) finding the overlap areas of the label classes in multidimensional (n-D) data, (b) evaluating ML models on these data, and (c) discovering boosted models on just the overlap areas. This paper is organized as follows: section 2 presents methodology and demonstration, It includes a subsection 2.1 on defining and finding pure and overlap areas in computational ML models, a subsection 2.2 on evaluating ML models on data in overlap areas, subsection 2.3 on the computational discovery of boosted models in overlap areas, and section 2.4 on the combined visual and computational discovery of models with a human-in-the-loop. Section 3 describes an open-source and free software system **JtabViz**, available on GitHub. The conclusion and future works are presented in section 4.

## 2    Methodology and Demonstration

### 2.1    Define and Find Pure and Overlap Areas in Computational ML Models

This section is devoted to (1) defining and finding the pure and overlap areas of classes, (2) generating synthetic cases in the overlap area with alternative definitions of the overlap area, and (3) generalizing the concept of the overlap areas for multiclass tasks.

**Defining and Finding Pure and overlap Areas for a Classification Model.** For different ML classification algorithms pure and overlap areas can significantly differ in both their definitions and contents for the same data, which we observe below.

- **Linear Classifiers** like Linear Discriminant Functions (LDF) and Logistic Regression (LR) have relatively similar definitions of the overlap areas with Support Vector Machine (SVM), and Neural Networks (NN), e.g., a Multi-Layer Perceptron (MLP), and many others. These are based on the unified concept of the interval [*a, b*] on the target attribute that we define below.
- **Tree Classifiers** like Decision trees (DT) do not have an equivalent of such an [*a, b*] interval on the target attribute. However, they do have misclassified cases associated with each leaf node. We treat some areas within attributes in leaf nodes as the overlap intervals. Thus, any tree classifier has several overlap intervals.
- **Ensemble Classifiers** like Random Forest (RF) these are classifiers with a voting mechanism, which require a specialized definition of the overlap interval.

Many ML algorithms can represent a **classifier model** discovered on an n-D dataset $D$ as a function $F(\mathbf{x})$, where $\mathbf{x}$ is an n-D point in D or a synthetic n-D point such that:

$$\text{If } F(\mathbf{x}) > T \text{ then } \mathbf{x} \in \text{class}_{\text{top}} \text{ else } \mathbf{x} \in \text{class}_{\text{bottom}} \tag{1}$$

where $T$ is the *threshold* positioned according to a selected separation strategy between these two classes by a given ML algorithm. Cases are misclassified when a labeled case is not on the correct side of the $F(\mathbf{x}) > T$ threshold for said class label.

We assume that for the ML classification model $F(\mathbf{x})$ its accuracy on the whole dataset $D$ and some splits of training and validation subsets are computed with cases misclassified by $F(\mathbf{x})$ are identified. We want to find the overlap area where all misclassified cases are located and the overlap interval in the $F(\mathbf{x})$ with values of misclassified cases. We define an **overlap interval** as $[a, b]$, for the ML classifier model defined as $F(\mathbf{x})$, as the *smallest interval* $[a, b]$ around threshold $T$, where $T \in [a, b]$, where all misclassified cases are located and

$$\text{for all } \mathbf{x} \in D \text{ if } F(\mathbf{x}) < a \text{ then } \mathbf{x} \in \text{class}_{\text{bottom}}, \text{ else if } F(\mathbf{x}) > b \text{ then } \mathbf{x} \in \text{class}_{\text{top}} \tag{2}$$

Thus, if the interval $[a, b]$ exists then all cases $\mathbf{x}$ from the training data $D$ such that $F(\mathbf{x})$ is less than $a$, i.e., below the overlap interval $[a, b]$, belong to class B (bottom class). Similarly, all cases $\mathbf{x}$ such that $F(\mathbf{x})$ is greater than $b$, i.e., above the interval $[a, b]$ belong to class U (upper class). Only cases $\mathbf{x}$ with $F(\mathbf{x})$ in the interval $[a, b]$, hence $a < F(x) < b$, can possibly be a mixture of cases of two classes including all misclassified cases. In other words, we have a *pure area* of class U, followed by an **overlap area** characterized by the **overlap interval** $[a, b]$, further followed by a *pure area* of class B. Moreover, we can define this overlap interval formally for a classifier $F(\mathbf{x})$.

**Definition.** An interval $[a, b]$ in called an **overlap interval** for a classifier $F(\mathbf{x})$ if

$$a = F(\mathbf{c}) = \min_{\mathbf{x} \in \text{Um}}(F(\mathbf{x})), \quad b = F(\mathbf{d}) = \max_{\mathbf{x} \in \text{Bm}}(F(\mathbf{y})) \tag{3}$$

where $\mathbf{c}$ is a case from a set Um of all misclassified cases of class U, and $\mathbf{d}$ is a case from a set Bm of all misclassified cases of class B from a selected dataset.

Fig. 1. illustrates this definition. Here case $\mathbf{c}$ is above the orange line as marked by the red class heatmap cell. It is the lowest case out of the Versicolor class (upper class), which is surrounded at the bottom by cases from the Virginica class (bottom class). Similarly, case $\mathbf{d}$ right below the upper orange line is marked cyan. It is the top-most case of the Virginica class (bottom class), which is at the top orange line and surrounded by cases from the Versicolor class (upper class). The selection of the dataset impacts the overlap interval $[a, b]$, which leads to the following data test. If the overlap interval found on the training data is much smaller than that found on the whole data, then the training data do not fully capture the n-D data distribution and are not fully representative to train the model $F(\mathbf{x})$, and we need to *modify the training data* to match intervals.

The overlap interval $[a, b]$ only characterizes the *overlap area* in n-D space but it does not describe it. Next we define the overlap area as a subset of the n-D space by building an **envelope** around all misclassified cases $D_m$, with the *top-most* n-D point of the envelope $\mathbf{e}_{\text{top}}$ as max values of all misclassified cases $D_m$, and the *bottom-most* n-D point of the envelope $\mathbf{e}_{\text{bot}}$ as min values of all misclassified cases $D_m$. Mathematically, two n-D points ($e_{bot}$, $e_{top}$) form a hyperrectangle/**hyperblock**/hyperbox (HB) [8, 11]. We will call this HB an **overlap area** for the ML classifier $F(x)$.

(a) Orange lines show top **d** and bottom **c** cases of the overlap interval. Threshold $T$ selected to minimize the error rate (strategy 1).

(b) After assigning the misclassified cases to their new classes with user chosen colors like the real class colors.

**Fig 1.** Illustration of the overlap interval definition with heatmap visualization and ordering cases by the value of a linear classifier $F(x)$.

**Definition.** A hyperblock with two n-D points ($e_{bot}$, $e_{top}$) forming its bottom and top edges is called an **overlap area** for the classifier $F(x)$, where $e_{top}$ is the n-D point formed from components of the max values of all misclassified cases $D_m$:

$$e_{top} = (max(x_1), max(x_2), \ldots, max(x_n)), \quad (4)$$

while $e_{bot}$ is the n-D point formed of the min values of all misclassified cases $D_m$:

$$e_{bot} = (min(x_1), min(x_2), \ldots, min(x_n)). \quad (5)$$

**Statement.** A hyperblock defined by two n-D points ($e_{bot}$, $e_{top}$) for the top and bottom edges is a *convex shape* (*bounding box*) in n-D space over a *convex hull* in n-D space built with all misclassified cases.

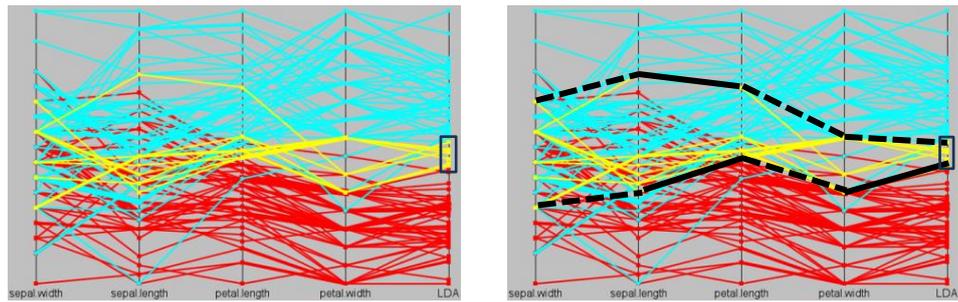

(a) Iris cases of two classes with yellow cases misclassified by linear classifier $F(x)$.

(b) Overlap area for a linear classifier $F(x)$ with black lines as a convex hull/envelope.

**Fig. 2.** Fisher Iris Versicolor and Virginica overlap are and convex hull for a linear classifier.

We can see it from the example with three points in 2D. They have a convex hull shaped as a tringle, and a bounding box as a rectangle surrounds this triangle. The advantage of using this *bounding-box as the overlap area* is in its *simple visualization* in the Parallel Coordinates in contrast with the *actual convex hull* over all misclassified cases in n-D space. This is illustrated in Figs. 2 and 3, which show the overlap area for a linear classifier and multilayer perceptron neural network on the Iris data. In Fig. 2 this area includes yellow polylines, which represent all cases misclassified by the model $F(\mathbf{x})$ from the dataset *D*. This overlap area is convex in n-D, but its representation in Fig. 2b is not convex. These cases have $F(\mathbf{x})$ values in the [*a*, *b*] *overlap interval* shown as a black box. The convex hull/envelope around these cases is shown by black lines, where solid lines are from actual misclassified cases and dotted lines are generated to form a convex hull.

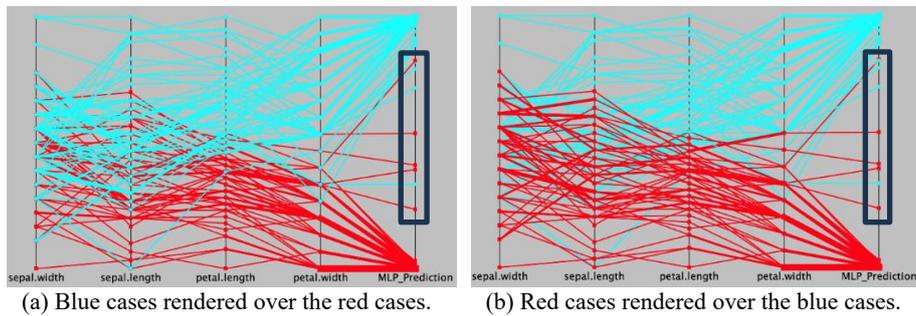

(a) Blue cases rendered over the red cases.   (b) Red cases rendered over the blue cases.
**Fig 3.** Predicting classes with a 4-hidden layer MLP neural network. Trained 1000 epochs, with loss of 2.06, learning rate 0.1. The black box is the overlap interval.

**Process of finding overlap intervals and areas.** To find the **overlap interval** [*a*, *b*] we search for two border cases *misclassified* by the ML classification model $F(\mathbf{x})$, we call them **c** and **d** above, which satisfy the mathematical condition (3). Figs. 1-3 not only illustrate the definition of the overlap interval and overall area, but also a human-navigated search process to produce them in the JTabViz software. Fig. 1 shows cases sorted by the value of $F(\mathbf{x})$ in decreasing order in the last column and using a heatmap to color cases, where a user can see the highest case **d** misclassified with a different color and the lowest misclassified case **c** also colored differently from nearby cases.

Fig. 2 illustrates the search process for the **overlap area** showing the resulting envelop surrounding all misclassified cases for the Iris data with a selected linear classifier $F(\mathbf{x})$. The steps to find it are: (1) find all misclassified cases by sorting data in descending order by the values of model $F(\mathbf{x})$, (2) compute top and bottom misclassified cases from them, (3) visualize all these cases in Parallel Coordinates.

**Generating synthetic cases in the overlap area with alternative definitions**. The motivation for generating new synthetic cases is that new real cases can differ from seen training cases. So, we *synthesize new cases* within the overlap area to explore the ML classifier model behavior. Therefore, we have several options like an even random distribution within the entire overlap HB, a Gaussian normal distribution around just the center n-D point of the overlap area HB, other representative distributions, or using methods like SMOTE and Generative Adversarial Networks (GANs).

Fig. 4 demonstrates the distribution-based approach. We use artificial slightly overlapping data, which have Gaussian distributions with slightly shifted distributions of 50 cases for 100 total cases thereby classified by an LDF. The black dotted line in Fig. 4a denotes the bottom edge of the pure area of the cyan class. Similarly, the black dotted line in Fig. 4b denotes upper edges of red class pure area.

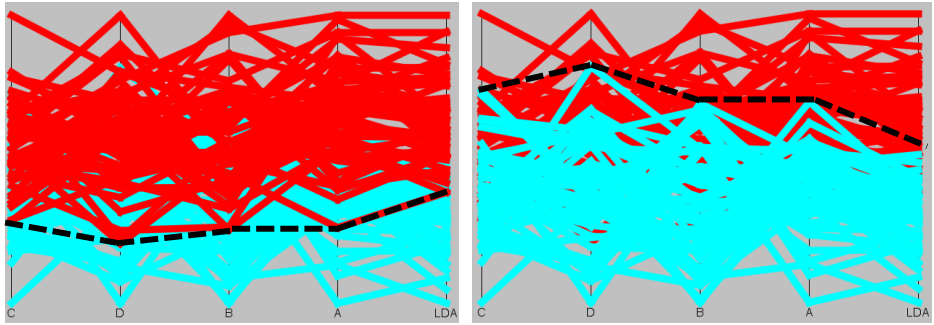

(a) Rendering artificial classes of red with higher LDA values on top of cyan.

(b) Rendering artificial cyan class on top with lower LDA values.

**Fig. 4.** Demonstrating the overlap definition on artificial slightly overlapping Gaussian distributions and classified by an LDA classifier. The black dotted lines denote the convex boundary between each class border, bottom then top respectively.

**Alternative definitions of the overlap area**. An HB between all min and max values of misclassified cases can *exaggerate the overlap area*. For instance, in Fig. 2b both top and bottom HB border edges are artificial, and respectively are misclassified by this LDF. We avoid such overgeneralization by modifying the envelop as shown in Fig. 5b. Fig. 5a shows an orange synthetic case. It goes through two edge segments of the overlap area, which are not segments of any real cases.

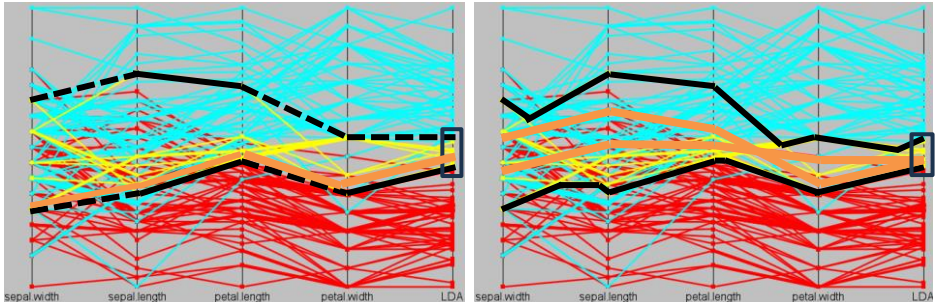

(a) Overgeneralized orange synthetic case follows two artificial dotted border edges of the overlap area. The black envelope around misclassified cases contains artificial edges (dotted lines) not from misclassified cases.

(b) Orange synthetic cases are fully within a modified envelope around misclassified cases. The black edges of the modified envelop belong to actual misclassified border cases.

**Fig. 5.** Comparison of two envelopes defining the overlap areas for a linear classifier $F(\mathbf{x})$ on the Fisher Iris Virginica (cyan) and Versicolor (red) classes. Cases in yellow are misclassified by $F(\mathbf{x})$ and found within the overlap interval $[a, b]$.

In contrast, Fig. 5b, which presents the modified envelope, does not allow the orange case shown in Fig. 5a. Here, the edges of the **modified envelope** follow only the lines of misclassified cases *avoiding overgeneralization* and *decreasing computational costs* by relaxing the envelope shape and overlap area size. It is unique being specific to the Parallel Coordinates visualization requiring finding the crossing points of inter-coordinate lines from different misclassified cases.

**Specifics of overlap area for a linear classifier**. A linear classifier model with *all positive coefficients* allows us to establish properties of the cases in the overlap area mathematically by using properties of a top case $\mathbf{e}_{top}$ and a bottom case $\mathbf{e}_{bot}$ from the envelope without generating and testing new synthetic cases in the envelope.

**Statement.** If $F$ is a linear classifier with all positive coefficients and both $\mathbf{e}_{top}$ and $\mathbf{e}_{bot}$ have values in the overlap interval $[a, b]$ of $F$, then for any case $\mathbf{x}$ in the envelope between $\mathbf{e}_{top}$ and $\mathbf{e}_{bot}$ the value of $F(\mathbf{x})$ also must be in the $[a, b]$ interval:

$$F(\mathbf{e}_{top}) \in [a, b] \,\&\, F(\mathbf{e}_{top}) \in [a, b] \,\&\, \mathbf{e}_{bot} < \mathbf{x} < \mathbf{e}_{top} \Rightarrow F(\mathbf{x}) \in [a, b]$$

**Proof**. With a property of n-D points that $\mathbf{e}_{bot} \leq \mathbf{x} \leq \mathbf{e}_{top}$ and all positive coefficients in $F$ we get $F(\mathbf{e}_{bot}) \leq F(\mathbf{x}) \leq F(\mathbf{e}_{top})$. Next, $F(\mathbf{e}_{top}) \in [a, b] \,\&\, F(\mathbf{e}_{top})$ gives $F(\mathbf{x}) \in [a, b]$.

Here we do not need to generate cases in the overlap area for such model $F(\mathbf{x})$ to check that $F(\mathbf{x})$ is in the overlap interval since it is mathematically provable. If a top and/or bottom case has $F(\mathbf{x})$ out of the overlap interval $[a, b]$ then we shrink $[a, b]$ to make new top and bottom cases to avoid exaggeration of the overlap area.

**Multiclass classification tasks**. Below we describe a process of defining and finding overlap areas and intervals for multiclass classification tasks. First, we get all misclassified cases for multiclass task from the classifier model $F(\mathbf{x})$. Next, we visualize all these cases with the Parallel Coordinates, and then we build a HB envelope around all of them. These steps are the same as with two classes. The last step of finding overlaps intervals in $F(\mathbf{x})$ differs. It is not a single overlap interval, however, for three classes we can get three pure intervals for these classes and intervals between them as overlap intervals. In general, it can be a set of $k$ overlap intervals $[a_1, b_1], [a_2, b_2], \ldots, [a_k, b_k]$ in the increasing order of their values. We start with finding the first interval at the bottom $[a_1, b_1]$ in the same way as for two-class models. This then continues to $[a_2, b_2]$ and other intervals sequentially until finding all these intervals in the same way.

**Model without a single function $F(\mathbf{x})$ and a single overlap interval**. For the models that do not have a single function $F(\mathbf{x})$ we do not generate overlap intervals on $F(\mathbf{x})$ as described above since they require a more specialized approach. For instance, a DT requires analysis of individual branching nodes, and their individual attributes to build overlap intervals within them. Some other classification models have two functions:

$$F_1(\mathbf{x}) > F_2(\mathbf{x}) \Rightarrow \mathbf{x} \in \text{Class 1, else } \mathbf{x} \in \text{Class 2} \qquad (6)$$

We convert these two functions to a new single function $F(\mathbf{x}) = F_1(\mathbf{x}) - F_2(\mathbf{x})$ and get a model with a single function $F(\mathbf{x})$. Hence, we generalize Eqs. (1) and (6) for Eq. (7):

$$F(\mathbf{x}) > 0 \Rightarrow \mathbf{x} \in \text{Class 1, else } \mathbf{x} \in \text{Class 2} \qquad (7)$$

The only difference of this function from Eq. (1) is that the threshold $T$ has a specific value of $T = 0$, with the process of finding the overlap interval $[a, b]$ being the same as described for Eq. (1). In summary of this section, we have found ways to identify the overlap areas, and overlap intervals, for ML classification models.

## 2.2. Evaluating ML Models on Data in Overlap Area

An ML model may get new cases that can be misclassified by a given classifier $F(\mathbf{x})$. How can we test the model before getting actual new cases? We propose using the *overlap interval* $[a, b]$ constructed for the model $F(\mathbf{x})$. If new synthetic cases $\mathbf{x}$ generated are outside of the overlap area for $F(\mathbf{x})$ have $F(\mathbf{x}) \notin [a, b]$, then it will be evidence that the areas outside of the overlap area are *pure*. This will add confidence to the model $F(\mathbf{x})$ outside of the overlap area and will justify the concentration of boosting the model on the found overlap area. Otherwise, it will be the evidence that both areas need to be boosted. See Fig. 6 with generated 25 cases from marginal distributions on each attribute in the pure area between yellow lines. It shows some synthetic cases, that were generated outside of the overlap n-D area, but are in the overlap interval, suggesting that more model boosting is needed.

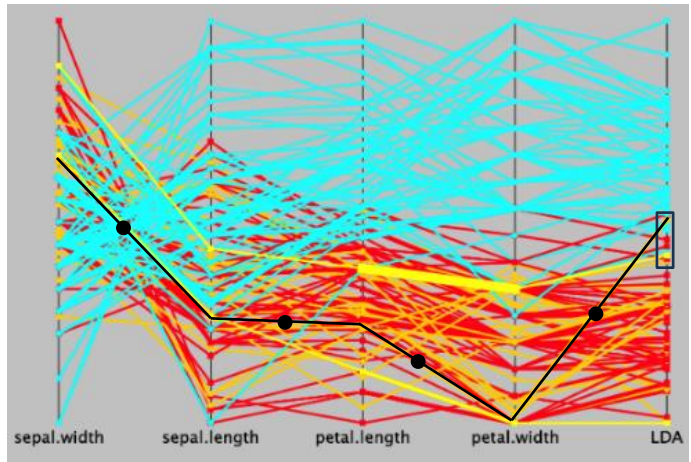

**Fig. 6.** Experiment with Iris data. A black synthetic case marked by black circles generated in the pure area but went to the overlap interval marked by the black rectangle. Orange lines are synthetic cases.

We evaluated the selected ML model in the overlap area by computing the *error rate* $E_{ovl} = N_{mis} / N_{ov}$ and the *accuracy* $A_{ovl} = 1 - N_{mis} / N_{ov}$, where $N_{ov}$ is the number of cases and $N_{mis}$ is the number of misclassified cases respectively in the envelope. A more detailed evaluation information provides a confusion matrix allowing analyzing performance for each class individually in the overlap area. The experiments show that the error rate computed for the overlap area is typically significantly higher than the error rate computed for the whole data [9]. Accuracy in the overlap area is often significantly lower than on the entire dataset, leading to the need of boosting pre-existing models and *discovering boosted models* on the overlap areas with better performances.

Another situation motivating the boosting of models on the overlap area is as follows. We may only get a small percent of cases (e.g. 10%) of all *real* cases $\mathbf{x}$ in the overlap area that end up in the overlap interval, $F(\mathbf{x}) \in [a, b]$ and potentially be misclassified by $F(\mathbf{x})$. In contrast, for *synthetic* cases we may get much greater percentage of such cases (e.g. 40%). This significant difference raises concern about the classifier $F(\mathbf{x})$ because it increases a risk of misclassification of new unseen real cases by $F(\mathbf{x})$.

This example demonstrates the need to evaluate models not only based on traditional accuracy and error rate but also on the *behavior relative to the overlap area.*

### 2.3. Computational Discovery of Boosted Models on the Overlap Area

In our experiment with LDF, presented in section 2.1 in Fig. 1, the misclassified cases constitute only 2 cases out of 100 total cases of the two Iris classes (2%). The overlap area contains 8 cases shown in Fig.1, which are 8% of all cases, and 2 misclassified cases that constitute 25% of the cases of the overlap area. These numbers are illuminating of the task difficulties, showing very different accuracies: high accuracy of 98% for all data, and a much lower accuracy of 75% for the more difficult overlap area. This difficult area contains only 8% of all data where we propose the efforts should be concentrated on developing a boosted model to reach the most desirable 100% accuracy.

Fig. 7 shows 100% accuracy reached by building a local LDF $F_2(\mathbf{x})$ only on the cases of the overlap area with coefficients $k_i$. The visualization is presented in In-Line Coordinates (ILC) defined in [7]. In Fig. 7 an orange line is the threshold line. All cases of the blue class are above the coordinate axis and all red class cases are below it. Each individual case $(x_1, x_2, x_3, x_4)$ is shown as a sequence of curves with points on the axis. The fist point has a distance from the origin point equal to $k_1 x_1$. The next point has a distance $k_1 x_1 + k_2 x_2$, similarly, the 3$^{rd}$ and 4$^{th}$ points have distances $k_1 x_1 + k_2 x_2 + k_3 x_3$ and $k_1 x_1 + k_2 x_2 + k_3 x_3 + k_4 x_4$, respectively. This final point represents a value of the linear classifier $F_2(\mathbf{x})$ on the overlap area. These last component points are visible next to the orange threshold line, with blue points on the left, and red points on the right.

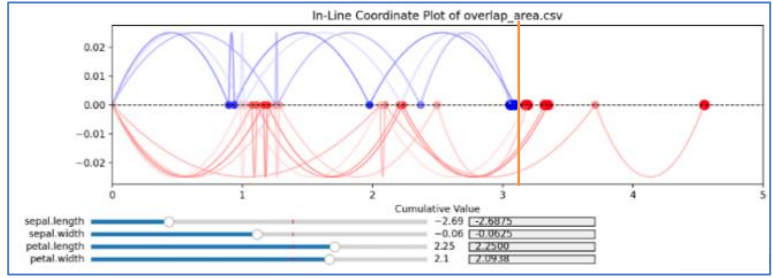

**Fig. 7**. Boosting process in the overlap area in ILC with automatic and interactive options.

Now we get a boosted model comprised of two classifiers $F_1(\mathbf{x})$ and $F_2(\mathbf{x})$ where $F_1(\mathbf{x})$ is $F(\mathbf{x})$ limited to $\mathbf{x}$ outside of the overlap area of $F(\mathbf{x})$, i.e., $F(\mathbf{x}) \notin [a, b]$ and $F_2(\mathbf{x})$ is a new classifier built for just the cases in the overlap area $[a, b]$ of $F(\mathbf{x})$, thus,

If $\mathbf{x} \notin [a, b]$ then classify $\mathbf{x}$ with $F_1(\mathbf{x})$, If $\mathbf{x} \in [a, b]$ then classify $\mathbf{x}$ with $F_2(\mathbf{x})$.     (8)

In contrast, in our experiment a Naïve Bayes classifier achieved only 62.5% accuracy on these overlap area cases. Boosting models like AdaBoost achieved perfect accuracy with 14 DT stumps with 4 parameters per stump (feature index, threshold, predicted node value, and weight) with total 56 parameters. In contrast our two linear functions are *much simpler* with only 10 parameters (four coefficients and one threshold for each linear function). Thus, we have 100 / 56 = 1.79 and 100 / 10 = 10 cases per parameter *avoiding overfitting* with two linear functions. In addition, CIVL allows to address class overlap using a computational-visual framework fully within the *interpretable class overlap boosting approach.*

Building a classifier $F_2$ depends on the task and the data. In the example above we used linear classifiers for $F_1$ and $F_2$. Below we present one of the methods to generate a boosted classifier which generalizes a traditional boosting approach. It assigns *two levels* of weights, where the largest weight $w_1$ is assigned to misclassified cases, and a lower weight $w_2$ is assigned to overlap area cases that are correctly classified. Respectively, the optimization criterion for discovering a boosted model is *not* the maximization of the accuracy, but rather, the maximization of the weighed accuracy:

$$A_w = w_1 N_{mis} + w_2 N_{cor\_Ov} \qquad (9)$$

This is for all overlap area training cases, where $N_{mis}$ is the number of misclassified cases in the overlap area, and $N_{cor\_Ov}$ is the number of correctly classified cases in the overlap area. Together models $F_1$ and $F_2$ will constitute a boosted model if $F_2$ outperforms the original model $F$ in the overlap area. To check that $F_2$ outperform $F$ on the overlap area, we compute and compare a standard accuracy of both in the overlap area.

### 2.4. Combined Visual/Computational Model Discovery with Human-in-the-Loop

This section continues expanding on the classical AdaBoost meta-algorithm, to improve accuracies of classification models from the analytical approach presented above. This is done by a combination of visual and computational tools with a human-in-the-loop. Whereas the classical boosting approach fundamentally relies on the base weaker computational classifiers to be strengthened, it prevents discovery of potentially more robust, accurate, and interpretable solutions. In contrast, the visual approach opens an opportunity for deeper data analysis before selecting base classifiers for boosting and developing classifiers by visual means.

**Finding pure and overlap areas with divide and classify process.** The model-based approach presented in previous sections requires the existence of an ML model already discovered, like a linear model, a DT, or a neural network. The proposed visualization-based approach *does not require* that an ML model already exists. It uses *only training data* to find pure and overlap areas in these data by combining *interactive lossless data visualization* (like Parallel Coordinates or other types of GLCs) and *computational means* to get a synergetic benefit. In this way we simultaneously discover an initial classification model $F$ and a boosted model $F_2$.

The example in Fig. 8 illustrates this process with *finding pure and overlap areas* in attributes for Iris data. A user equipped with interactive JTabViz tools that we developed (see section 3) marks pure areas in each attribute. Fig. 8a shows some pure areas marked by a user (see yellow rectangles). The remaining cases are shown on the Fig. 8b, which are candidates to be in the overlap area. A user can refine interactively this dataset further with removal another pure area marked in yellow in Fig. 8b. A user in this example, can not only easily find the pure areas interactively, but also can get them *automatically* by using functionality of the JTabViz software. Discovering rectangles directly leads to the *interpretable classification model $F_1$* on pure areas with 100% accuracy in the form of a set of *classification logical propositional rules* like:

If $0.75 \leq$ petal.width($\mathbf{x}$) $\leq 1.00$ then $\mathbf{x}$ belongs to Class Virginica (Count: 34 cases, 68% of class, 34% of the total dataset).

Moreover, in this example, a user can easily notice that some yellow rectangles are *redundant* and not needed to build rules to cover all cases that are in the pure areas. For instance, just one of the yellow rectangles, 1 and 2, on the last two coordinates in Fig. 8a is sufficient to discriminate the red class (Setosa). Next, a user can search and remove redundant rectangles interactively in JTabViz software or use a computational tool in JTabViz to remove redundant rectangles automatically and build and output a smaller set of rules.

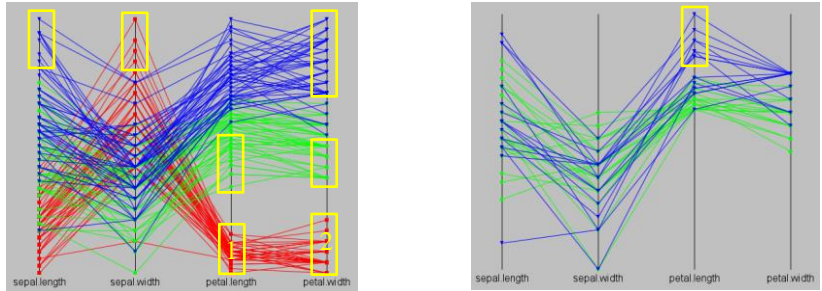

(a) Full data with pure areas marked by yellow rectangles

(b) Remaing cases after removal cases in the marked pure araes.

**Fig. 8**. Min-max normalized Iris data in Parallel Coordinates: (a) full data, (b) overlap data.

The combination of computational and interactive visual tools in JTabViz is especially beneficial for complex datasets. A user interactively builds rectangles for pure areas, and backend computational tools sort pure regions by coverage size, extracting exact values of the limits of the rectangles to generate exact rules. The produced model is a *partial model*, as it does not classify cases outside of the discovered pure areas. The advantage of the described interactive process is that the user can opt to skip some pure areas, considering them as too small of spatial size or case coverage, which can lead to *overfitted models/rules* [6]. These cases will be in the set of remaining cases $R$, which includes the set of actual overlap cases $O$. Next a user works on attempting to discover a model $F_2$ on the remaining cases R, which contains all overlap cases. This process differs from the process proposed in section 2.1 - 2.3, where we start with the model $F$ discovered by a traditional automatic ML algorithm and then discover a model $F_2$ as its boosted version on the overlap area. Now, we have a partial model $F_1$ that is defined on the pure areas, and we do not have model $F_2$ at all. We need to build $F_2$ from scratch of the remaining cases.

Here we propose two options: (1) building a traditional ML model $F_2$ like LDF, DT or neural network computationally on the remaining cases $R$, and (2) building model $F_2$ using a combination of visual and computational tools. The option (1) with traditional ML algorithms brings us to the same approach as we proposed for the whole dataset in sections 2.1 - 2.3 to build a first model for the remaining cases and to boost it as we described in sections 2.1 - 2.3.

The option (2) with a combination of visual and computational tools continues the same process that we described above. In short it is a **divide and classify** approach where we search for pure areas for the remaining cases until all cases will be classified. At first glance it looks impossible to continue the process in the overlap area because all pure areas already have been found. However, after removal all cases of the pure areas, new pure areas can emerge in the remaining cases only which happened on real data, as we show in the next section, with an example in the same Iris data. An advantage of this approach is that it guarantees a classification of all given training cases with full interpretability. The disadvantage of this approach is that in the full extension it will lead to overfitting, and a user would need to stop this process to avoid overfitting. A user can follow a practice known as pruning for DTs, which joins nodes with a small number of cases leading to decreased accuracy. Here a user can build an envelope over small pure areas to get larger ones allowing a decreased accuracy [5].

**Iterative divide and classify process.** Below we present a divide and classify method using the same Iris data example that we used in Figs. 1, 2 and 4. We already demonstrated there that we can build a simple rule with pure areas in Fig. 8a to classify accurately all cases of class Setosa. Therefore, we concentrate here on the classification of the cases of remaining two classes Virginia and Versicolor.

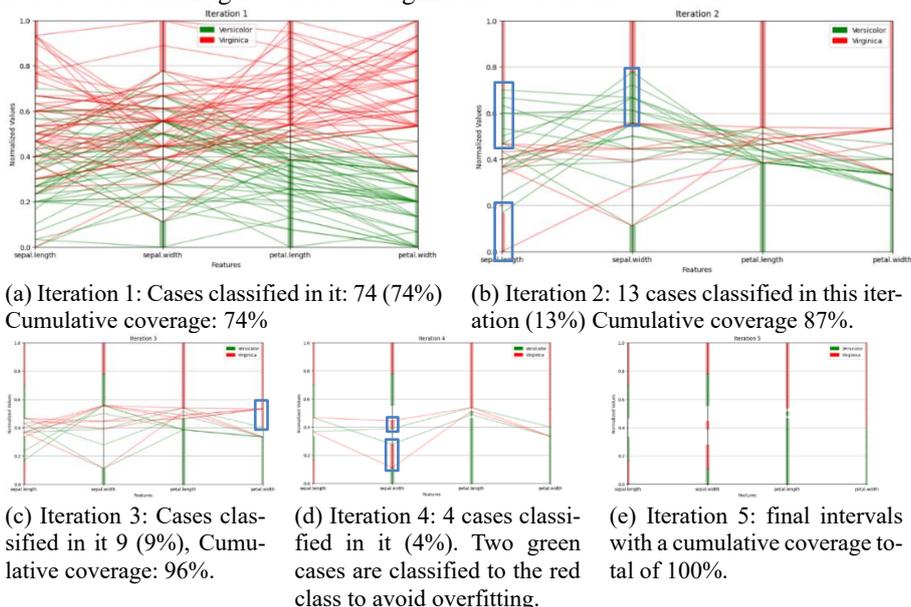

(a) Iteration 1: Cases classified in it: 74 (74%) Cumulative coverage: 74%

(b) Iteration 2: 13 cases classified in this iteration (13%) Cumulative coverage 87%.

(c) Iteration 3: Cases classified in it 9 (9%), Cumulative coverage: 96%.

(d) Iteration 4: 4 cases classified in it (4%). Two green cases are classified to the red class to avoid overfitting.

(e) Iteration 5: final intervals with a cumulative coverage total of 100%.

**Fig. 9.** Classification of Iris Virginica and Versicolor with single attribute rules in the Divide and Classify iterative process.

Fig. 9a shows all cases of these two classes along with the results of the first iteration of the divide and concur process. These results are represented by thick red and green lines on each coordinate, which capture pure area in coordinates. A user can capture them interactively or computationally with JTabViz software and adjust interactively. Next a user generates the rules from them with this software. Then a user removes all cases from the found pure areas as shown in Fig. 9b. The cases on Fig. 9b are a source of the second iteration of finding new pure areas. It shows found new pure areas, which are marked by the blue rectangles. These areas have not been visible in Fig. 9a because some removed cases have points in these areas. The process continues as shown in Figs. 9c and 9d to iterations 3 and 4. Iteration 3 removed 9 cases by creating a small pure area shown there as a blue rectangle. Iteration 4 needs to deal with 4 remaining cases shown on the left of Fig. 9d. A user joins two green cases to the red class as shows by two blue rectangles to avoid further overfitting.

In this process thresholds are set, and adjusted, as determined by the subject expert, while the JTabViz software system automatically searches for single-attribute rules, which have coverage greater than the set threshold value. Thresholds set up without user guidance can results in too many rules, therefore, a human-in-the-loop can filter out inefficient rules. As we see, a user analyzes the obtained rules, joins or reverses some of them, considering them too small, because they cover only few cases or only in very small intervals. Other boosting options can also be available for the user. We discussed these options later in this paper by developing new types of attributes. This example demonstrated our general statement that we have a dual process here of both visual and computational means assisting in building better boosted models.

The advantage of this divide and classify process is that a user who is a domain expert is in full control of the model development, including control over overfitting. A user can decide when rules need to be combined to avoid overfitting. The model is fully interpretable, and its discovery does not require, from the user, any sophisticated ML knowledge to understand it. Below we present a *general form of singe-attribute rules*: Let $\{Int_{ij}\}$ is a set of pure intervals of iteration $i$. If case $\mathbf{x}$ has an attribute in the interval $Int_{ij}$ then we will write $Int_{ij}(\mathbf{x})$ = true. Each interval produces a pure rule $R_{ij}$.

$R_{ij}$: If $Int_{ij}(\mathbf{x})$ = true then $L$, where $L$ is class label, e.g., $L$ can be 1,2 or 3.

In other words, each rule R has three arguments: case $\mathbf{x}$, interval Int, and label $L$.

$R(\mathbf{x}, Int, L)$ = true if $Int(\mathbf{x})$ = true and $L$ is a label that is assigned to the interval Int.

For instance, we can get the following rules at iteration one,

Iteration 1: $R_{11}$: If $Int_{11}(\mathbf{x})$ = true then $L = 1$, $R_{12}$: If $Int_{12}(\mathbf{x})$ = true then $L = 2$, and $R_{13}$: If $Int_{13}(\mathbf{x})$ = true then $L = 3$.

Due to the iteration, rules for each subsequent iteration depend directly on the prior iterations. Therefore, we cannot write them down as we did for iteration one. Rules from each iteration requires that the rules from the previous iterations will be false (not applicable) for the cases in the next iteration. Thus, the rules for iteration two are:

$R_{21}$: If $Int_{21}(\mathbf{x})$ = true & (not $R_{11}$) & (not $R_{12}$) & (not $R_{13}$) then $L = 1$,
$R_{22}$: If $Int_{22}(\mathbf{x})$ = true & (not $R_{11}$) & (not $R_{12}$) & (not $R_{13}$) then $L = 2$,
$R_{23}$: If $Int_{23}(\mathbf{x})$ = true & (not $R_{11}$) & (not $R_{12}$) & (not $R_{13}$) then $L = 3$.

Similarly, the rules from iteration three are:

$R_{21}$: If $Int_{21}(\mathbf{x})$ = true & (not $R_{11}$) & (not $R_{12}$) & (not $R_{13}$) & (not $R_{21}$) & (not $R_{22}$) & (not $R_{23}$) then $L = 1$,
$R_{22}$: If $Int_{22}(\mathbf{x})$ = true & (not $R_{11}$) & (not $R_{12}$) & (not $R_{13}$) & (not $R_{21}$) & (not $R_{22}$) & (not $R_{23}$) then $L = 2$,
$R_{23}$: If $Int_{23}(\mathbf{x})$ = true & (not $R_{11}$) & (not $R_{12}$) & (not $R_{13}$) & (not $R_{21}$) & (not $R_{22}$) & (not $R_{23}$) then $L = 3$.

While the mathematical form of these rules looks complicated, the visual form of these rules, and the relations between them, is simple and demonstrates the recursive rule structure visualized in Fig. 10a, which again does not require the user to have any sophisticated mathematical knowledge of ML.

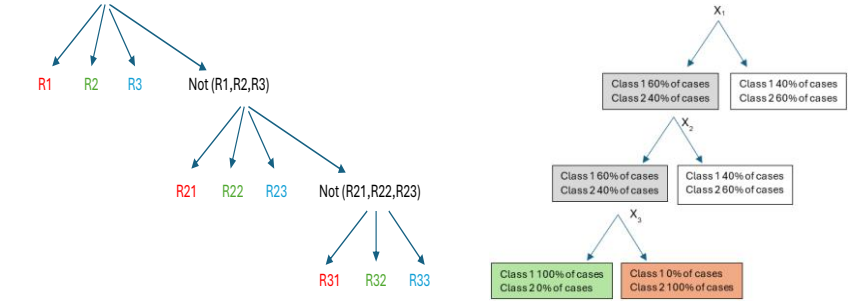

(a) Generalized DT visualizes relations between pure nodes (rules generated iteratively).  (b) Traditional binary DT has impure nodes with single threshold split of attributes.

**Fig. 10.** Comparison of a generalized DT (a) and traditional binary DT (b).

The visual form of these rules is much simpler to understand, which are easier to interpret than with analytical discovery. Also, these rules can be easier to understand than traditional DTs for the user. To see this, consider a DT in Fig. 10b, where attribute $X_1$ at the root node splits cases to two classes at 60% and 40% of cases in each class to the children nodes, this is close to a random split of 50:50. The next attribute $X_2$ at the next layer splits the remaining cases in the same 60%: 40% ratio, and only attribute $X_3$ after these initial layers splits cases to two classes purely (100% to 0).

What can a user do to interpret this branch of the DT meaningfully? A user visually can see that the first three attributes are only marginally able to separate classes 1 and 2 on the left branch, and only the very last attribute $X_3$ finally was able to classify all remaining cases correctly. It means that for the user the first three attributes have a very minimal meaningful value because they do not significantly separate classes, rather they preprocess the data for the final attribute $X_3$. In contrast all rules in Fig. 10a produced by the divide and classify process are highly informative for the user, because all of them separate classes. Next, the nodes in Fig. 10a are based on the intervals on the attributes, not a single threshold on the attribute as traditional DTs in Fig. 10b. Thus, the proposed process produces a **generalized non-binary DT** with rules as nodes that can be *more informative* than a traditional DT for the user.

**Boosting models with FoL rules beyond a single attribute: monotonic property**. Traditional DTs and generalized DTs that we discussed above exploit a *single* attribute. In the traditional DTs, it is like $x_i \geq T$. In the generalized DTs it is like $T_1 \geq x_i \geq T_2$, i.e., $x_i \in [T_1, T_2]$. While proposed above *generalized non-binary DT models* generalize the traditional DTs, they are still equivalent to propositional rules with unary predicates, $P(x_i) = \text{true} \Leftrightarrow x_i \geq T$ or $P(x_i) = \text{true} \Leftrightarrow T_1 \geq x_i \geq T_2$ with one variable $x_i$. Below we propose boosting ML models with **First Order Logic (FoL)** *rules/models,* which are more general and richer than propositional rules with a single variable. For this, we discover monotonic relations between attributes in this section, like $x_1 > x_2 > x_3 > x_4$, which has FoL form of a predicate $P(x_1, x_2, x_3, x_4) = \text{true} \Leftrightarrow x_1 > x_2 > x_3 > x_4$, when we discover a monotonic relation between four attributes. When such monotonic property $P(x_1, x_2, x_3, x_4)$ is discovered it can be converted to the **FoL classification rule**:

If $\mathbf{x} = (x_1, x_2, x_3, x_4)$ & $P(x_1, x_2, x_3, x_4)$ = true then $\mathbf{x} \in$ Class 1, else $\mathbf{x} \in$ Class 2   (10)

An example where this property was discovered by visual analysis of data in Parallel Coordinates is on the same Iris data in their original non-normalized form. This property is present only for Setosa class allowing to separate it from two other classes:

If $\mathbf{x} = (x_1, x_2, x_3, x_4)$ & $P(x_1, x_2, x_3, x_4)$ = true then $\mathbf{x} \in$ Setosa class,
else $\mathbf{x} \in$ Virginica or Versicolor classes   (11)

Note that, normalization of all attributed to [0, 1] interval destroys the original monotonic property, which is present in the original data. This points out on necessity to analyze not only normalized data, which is commonly used in parallel coordinates, but in the original data form too. Specifically, all cases of the Setosa class have the monotonic property in the original non-normalized data, values in centimeters:

$$Sepal\ Length > Sepal\ Width > Petal\ Length > Petal\ Width \quad (12)$$

This monotone decreasing property is visualized in Parallel Coordinates in Fig. 8a, which was confirmed computationally. This visually discovered monotonicity property leads to the **monotonic FoL rule** $R_M$:

If $Sepal\_Length(\mathbf{x}) > Sepal\_Width(\mathbf{x}) > Petal\_Length(\mathbf{x}) > Petal\_Width(\mathbf{x})$

$$\Longrightarrow \mathbf{x} = (x_1, x_2, x_3, x_4) \in \text{Setosa class} \quad (13)$$

At first glance, this FoL rule is deficient, being more complex than two propositional rules with a single attribute each, which are also visually discoverable by observing Setosa class in Fig. 11:

$R_{PL}$: If $Petal\ Length(\mathbf{x}) < T_{PL}$ then $\mathbf{x} = (x_1, x_2, x_3, x_4) \in$ Setosa   (14)

$R_{PW}$: If $Petal\ Width(\mathbf{x}) < T_{PW}$ then $\mathbf{x} = (x_1, x_2, x_3, x_4) \in$ Setosa   (15)

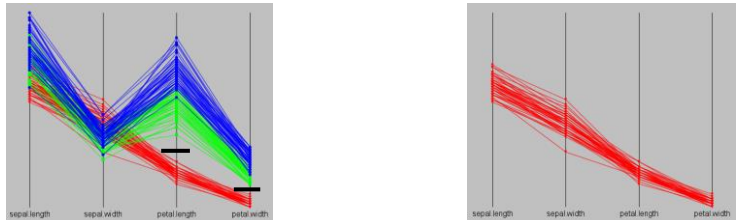

(a) Visualizing Setosa class monotonicity.     (b) Visualizing the Setosa class by itself.
**Fig. 11.** Iris 4-D data in parallel coordinates with a visual downward monotonic trend of Setosa class with attribute order of: sepal length → sepal width → petal length → petal width; Setosa (red), Virginica (blue), and Versicolor (green).

Fig. 11a shows $T_{PL}$ and $T_{PW}$ decision thresholds marked by short black lines for these propositional rules. These rules are discoverable by DTs, but they have an uncertainty deficiency. Different numerical values of thresholds can satisfy Eqs. (15) and (16).

Multiple such threshold values separate training data equally well but can classify new cases differently. In contrast the monotonic rule (13) is more robust being free from this threshold uncertainty problem not having the threshold. Next, while simple rules (14) and (15) were discovered in these data, it is hard to expect that such simple rules will be common for more complex data. Therefore, discovering monotonic rule can be a valuable boosting alternative in many datasets.

Another advantage of the monotonic rule (13) over single-attribute rules (14) and (15) is that it captures all four properties. A single threshold $x_i < T$ does not restrict values of other attributes allowing their overgeneralization to any unrealistic values.

**Generalized Divide and Classify approach.** The consideration above of monotonic property allows us to formulate the divide and classify approach more generally. It starts with a dataset $D$ containing all cases to be classified, where each case has $n$ attributes, or features. Then it identifies a subset $P \subset D$ using FoL rules like FoL monotonic rule (13). A subset $P$ consists of cases from a single class that fall into pure, easily classifiable n-D regions with high classification confidence. The remaining set is now $D' \coloneqq D \setminus P$. This process is iterated until no cases remain in $D$. At each iteration a user searches for a distinct first order logic property that can be converted to a classification rule. In addition, the user involvement in each iteration allows ensuring that individual cases of high importance can get attention beyond overall accuracy of the model.

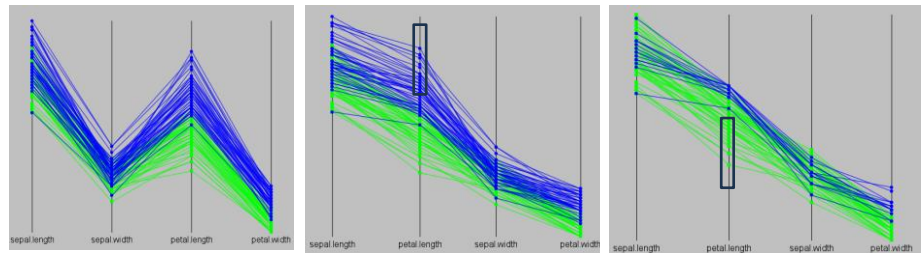

(a) Removed the Setosa class with remaining classes visualized allowing for a new focus on previously occluded areas.

(b) Swapping sepal width and petal length attributes reveals patterns like if *Petal Length* > 5.1 then 30 of 50 cases are Virignica.

(c) Removed cases with *Petal Length* > 5.1. Another pure green area in the black rectangle is ready to be removed to reveal the overlap area.

**Fig. 9.** Divide and Classify process using newly revealed patterns on Iris data after removal of entire Setosa class in parallel coordinates.

This iteration process is illustrated in Fig. 9. In the Iris example, the monotonic rule (13) classifies all 50 cases of the Setosa class, so $D'$ is $D$ without any Setosa cases as shown in Fig. 8. After hiding the Setosa subset of $D$, we can see the data shown in Fig.9a of two remaining classes, but without any monotone property visible. Next, we reorder attributed as shown in Fig. 9b, which reveals the monotone decrease of attribute values in both classes. Thus, while monotonic property is discovered, but it is not a discriminative feature of these classes. However, monotonic visualization in Fig. 9b simplifies the visual analysis of data including finding pure areas in individual attributes, like Petal Length > 5.1 similarly to the process used in Fig 2. The result of this iteration is shown in Fig. 9c. A user can continue a process of separating cases of these two classes by analyzing Fig. 9c. For instance, a user can continue to extract pure areas on a single attribute like shown in the black box in Fig. 9c.

In addition, or alternatively, a user can continue boosting and developing the model by noticing the difference in slopes between attributes for blue and green cases. It seems that in Fig. 9c, the slope of green cases **u** are *greater than* slopes of the blue cases **w** for the first two attributes, which we will denote as $x_1$ and $x_2$. This can be expressed by a FoL property: $x_1(\mathbf{u}) - x_2(\mathbf{u}) \geq x_1(\mathbf{w}) - x_2(\mathbf{w})$.

A user can also notice an opposite property for slopes between attributes $x_2$ and $x_3$: $[x_2(\mathbf{u}) - x_3(\mathbf{u}) \leq x_2(\mathbf{w}) - x_3(\mathbf{w})]$. A user can continue this observation and suggest a relation like: $[x_3(\mathbf{u}) - x_4(\mathbf{u}) \geq x_3(\mathbf{w}) - x_3(\mathbf{w})]$. This visual observation allows generating hypothetical FoL rules that can be tested computationally on their accuracy:

$$\text{If } \mathbf{u} \in \text{green class} \ \& \ [x_1(\mathbf{u}) - x_2(\mathbf{u}) \geq x_1(\mathbf{w}) - x_2(\mathbf{w})] \text{ then } \mathbf{w} \in \text{blue class} \tag{16}$$

$$\text{If } \mathbf{u} \in \text{green class} \ \& \ [x_1(\mathbf{u}) - x_2(\mathbf{u}) \geq x_1(\mathbf{w}) - x_2(\mathbf{w})] \ \& \ [x_2(\mathbf{u}) - x_3(\mathbf{u}) \leq x_2(\mathbf{w}) - x_3(\mathbf{w})]$$
$$\text{then } \mathbf{w} \in \text{blue class} \tag{17}$$

$$\text{If } \mathbf{u} \in \text{green class} \ \& \ [x_1(\mathbf{u}) - x_2(\mathbf{u}) \geq x_1(\mathbf{w}) - x_2(\mathbf{w})] \ \& \ [x_2(\mathbf{u}) - x_3(\mathbf{u}) \leq x_2(\mathbf{w}) - x_3(\mathbf{w})]$$
$$\& \ [x_3(\mathbf{u}) - x_4(\mathbf{u}) \geq x_3(\mathbf{w}) - x_3(\mathbf{w})] \text{ then } \mathbf{w} \in \text{blue class} \tag{18}$$

These are just examples of FoL rules that can be generated by visual data exploration and then tested computationally. With this process the subject matter expert is computationally assisted and advised to explore multiple rule constructions. Multiple rulesets may identify the same areas with differing accuracy and interpretability since class boundaries are naturally fuzzy, with imbalanced case counts typically. This interactive process is conducted in a human-driven search of sufficient classifiable patterns for their specific data domain of expertise. Development of both computational and visual methods would further benefit this process.

**Feature Engineering.** The method proposed above to boost ML models with FoL rules by discovering properties that involve several attributes can be expanded to feature engineering. Inspired by FoL properties like $[x_1(\mathbf{u}) - x_2(\mathbf{u}) \geq x_1(\mathbf{w}) - x_2(\mathbf{w})]$, we generate new features/attributes like $x_5 = x_1 - x_2$, $x_6 = x_2 - x_3$, and $x_7 = x_3 - x_4$, and use them with the original attributes in the **traditional ML algorithms** to discover more accurate models.

With the right engineered features, the mistakes made by ML classifier algorithms can be reduced. Within the JtabViz system we provide software tools to build new features, explore, and visualize them. In the Iris example new features as differences between length and width of petal and sepal of flowers are meaningful interpretable features, but not all engineered features can be interpretable in the domain requiring extra effort to make features interpretable.

Therefore, in general, discovery of the right features requires a domain familiarity, i.e. subject matter expertise with a human-in-the-loop. Feature engineering and orderings by their correlation help *simplify the overlap area* of classes and its visualization. Software system JtabViz

The **JTabViz: Java Tabular Visualization Toolkit** is a software tool developed for this research project on boosting classifier algorithms and visual representation learning. JTabViz supports ML data analysis, classification, and lossless data visualization. It is fully open-source and freely available under the MIT license at the CWU-VKD-LAB Github: *https://github.com/CWU-VKD-LAB*. It supports the following ML algorithms: DT, k-NN, LDA, PCA, RF, MLP, and Support Sum Machine (SSM), a novel multi-class optimization-based linear type LDA. Users load tabular data to visualize and classify. Several GLC visualization methods are supported including Parallel, In-Line, Circular, and other coordinates. Other visualization methods include tabular heatmaps, and covariance matrix heatmaps. Additional functionalities include row manipulation, classification rule testing with confusion matrices, automatic pure region discovery, attribute covariance sorting, and feature engineering where options include computing new features from existing ones as slopes, distances, weighted sums, forward/backward differences, and trigonometric function values.

## 4. Conclusion and future work

We proposed a **Computational and Interactive Visual Learning** (**CIVL**) framework with a Human-in-the-Loop as a new way to boost and discover classification models. The idea of the CIVL framework is to separate areas of feature space to areas where models can be discovered easier than in other areas, where more effort is needed to discover a model, and where the *misclassified cases* are located. The proposed **Divide and Classify (D&C)** strategy, or a *Separating Scheme*, consists of searching for pure and overlap areas of classes, and building separate models in these areas. We defined and demonstrated the CIVL framework with computational ML algorithms and models, like linear and nonlinear classifiers, generalized DTs, logical propositional and first order rules, and neural networks. The implemented visual learning part is based on the Parallel and In-Line Coordinates and can be expanded to other GLCs in future work.

We showed the benefits of this framework for improving accuracy and explainability of ML models by allowing (1) decreasing significantly the amount of data requiring complex exploration for model discovery, (2) avoiding inflated ML model accuracy estimates, which is critical for high-stakes ML tasks, (3) getting accurate interpretable propositional and First order Logic rules, and (4) involving domain experts in model boosting and discovery interactively in lossless n-D data visualization space directly.

This work demonstrated a linear 100% accurate Fisher Iris model which is much simpler than neural networks, DTs, and traditional boosting models with a much fewer number of parameters. We have already conducted multiple experiments on larger datasets like MNIST which we cannot include here due to the paper size limitations, these will be reported on in the future.

In summary, focusing on just overlap areas is a promising way to further boosting and improve ML models, leading to major progress. Dealing with the overlap areas without lossless visualization is difficult or impossible for domain experts, therefore, Visual Knowledge Discovery methods need to be expanded. Future work includes, (1) exploring the proposed framework on larger datasets with more cases and features, (2) expanding the methods implemented in the framework, and (3) expanding the proposed computational and visualization framework to other types of ML algorithms and visualization methods.